\documentclass[runningheads]{llncs}
\usepackage[T1]{fontenc}
\usepackage{graphicx}
\usepackage{booktabs}
\usepackage{multirow}
\usepackage{url}
\usepackage{amsmath} 
\usepackage{subcaption}
\usepackage{orcidlink}

\renewcommand{\orcidID}[1]{\orcidlink{#1}}

\begin{document}
\title{Evaluating the Sensitivity of BiLSTM Forecasting Models to Sequence Length and Input Noise}
%
\titlerunning{Sensitivity of BiLSTM Forecasting to Sequence Length and Noise}
\author{Salma Albelali\inst{1,2}\orcidID{0000-0002-2819-5511} \and
Moataz Ahmed\inst{1,3}\orcidID{0000-0003-0042-8819}}
\authorrunning{S. Albelali and M. Ahmed}
%
\institute{King Fahd University of Petroleum \& Minerals, Department of Information and Computer Science, Dhahran, Saudi Arabia \and
Imam Abdulrahman Bin Faisal University, Department of Computer Science\\
\email{Salbelali@iau.edu.sa} \and
SDAIA-KFUPM Joint Research Center for Artificial Intelligence, Dhahran, Saudi Arabia\\
\email{Moataz@kfupm.edu.sa}}

\maketitle              

\begin{abstract}

Deep learning (DL) models, a specialized class of multilayer neural networks, have become central to time-series forecasting in critical domains such as environmental monitoring and the Internet of Things (IoT). Among these, Bidirectional Long Short-Term Memory (BiLSTM) architectures are particularly effective in capturing complex temporal dependencies. However, the robustness and generalization of such models are highly sensitive to input data characteristics—an aspect that remains underexplored in existing literature. This study presents a systematic empirical analysis of two key data-centric factors: input sequence length and additive noise. To support this investigation, a modular and reproducible forecasting pipeline is developed, incorporating standardized preprocessing, sequence generation, model training, validation, and evaluation. Controlled experiments are conducted on three real-world datasets with varying sampling frequencies to assess BiLSTM performance under different input conditions. The results yield three key findings: (1) longer input sequences significantly increase the risk of overfitting and data leakage, particularly in data-constrained environments; (2) additive noise consistently degrades predictive accuracy across sampling frequencies; and (3) the simultaneous presence of both factors results in the most substantial decline in model stability. While datasets with higher observation frequencies exhibit greater robustness, they remain vulnerable when both input challenges are present. These findings highlight important limitations in current DL-based forecasting pipelines and underscore the need for data-aware design strategies. This work contributes to a deeper understanding of DL model behavior in dynamic time-series environments and provides practical insights for developing more reliable and generalizable forecasting systems.

\keywords{Neural Network Models \and Deep Learning \and Data Quality}
\end{abstract}

\section{Introduction}

Deep learning (DL) models have become indispensable in modern predictive analytics, supporting applications across diverse domains such as finance, healthcare, energy systems, and the Internet of Things (IoT). Within this landscape, recurrent neural networks—particularly Bidirectional Long Short-Term Memory (BiLSTM) networks—have demonstrated strong performance in time-series forecasting by capturing complex and long-range temporal dependencies. By processing input sequences in both forward and backward directions, BiLSTM architectures extract rich contextual information that is often crucial for accurate prediction.

Despite their empirical success, the performance and reliability of BiLSTM models are highly sensitive to characteristics of the input data. In time-series forecasting, two key factors—\textit{sequence length} and \textit{input noise}—have a significant impact on model generalization and stability. Sequence length determines the temporal window used for prediction. While longer sequences can expose richer historical patterns, they may also lead to overfitting, particularly in low-resolution or data-scarce settings such as daily climate records. In contrast, shorter sequences may result in underfitting due to inadequate temporal context. Input noise presents a different but equally important challenge. Noise can arise from a variety of sources, including sensor errors, transmission delays, or environmental fluctuations. Even minor perturbations can distort underlying patterns and degrade forecasting performance. This is especially critical in high-stakes environments where small prediction errors can compound over time, leading to costly or unsafe outcomes.

Overfitting and limited generalization remain central challenges in DL-based forecasting, particularly under shifting input conditions. Models that overfit are prone to memorizing training data \cite{hui2021privacy}, while poor generalization has been linked to unstable or inconsistent behavior across unseen inputs \cite{chen2020machine,yeom2018privacy}. These vulnerabilities not only impair predictive reliability but also raise concerns for responsible deployment in data-sensitive domains. While prior studies have explored the individual effects of sequence length and noise, their combined influence on BiLSTM model robustness remains insufficiently examined. To address this gap, we present a systematic empirical investigation into how sequence length and additive Gaussian noise jointly affect the robustness, generalization, and stability of BiLSTM-based time-series forecasting models. Controlled experiments are conducted across three real-world weather datasets, each representing different sampling frequencies and temporal resolutions, to evaluate model behavior under a variety of input conditions.

This study adopts a data-oriented testing framework to analyze the sensitivity of BiLSTM models to variations in sequence configuration and input noise. Our results demonstrate that these factors significantly influence model performance and highlight trade-offs between predictive accuracy, overfitting, and robustness. The findings offer practical insights for developing more resilient time-series forecasting pipelines based on deep learning.

The primary objectives of this study are as follows:

\begin{itemize}
    \item Assess the impact of sequence length and noise injection on the performance and robustness of BiLSTM-based time-series forecasting models.
    \item Evaluate how noise injection influences model stability and generalization under varying data conditions.
    \item Design testing methodologies to identify data and model configurations that affect forecasting reliability.
\end{itemize}

This study provides the following key contributions:

\begin{itemize}
    \item An empirical evaluation of BiLSTM behavior under different input configurations, analyzing the effect of sequence length and noise injection on model robustness and generalization.
    \item A systematic investigation of how noise injection impacts predictive stability across datasets, including its trade-offs with accuracy and overfitting.
    \item A multi-dataset experimental analysis comparing the effects of long-sequence modeling across high-frequency (hourly) and low-frequency (daily) time-series data.
\end{itemize}

The remainder of this paper is organized as follows. Section~\ref{sec:related_work} reviews prior work related to sequence length, noise injection, and their implications for model robustness and reliability in time-series forecasting. Section~\ref{sec:methodology} describes the proposed methodology, including dataset preprocessing, model architecture, experimental design, and evaluation metrics. Section~\ref{sec:results} presents the experimental results and examines the impact of sequence length and noise on model behavior. Section~\ref{sec:future_work} discusses potential directions for future research. Finally, Section~\ref{sec:conclusion} summarizes the key findings and concludes the paper.

\section{Literature Review}
\label{sec:related_work}

The reliability of DL models such as BiLSTM is strongly influenced by data-centric factors like sequence length and input noise. Sequence length is a fundamental configuration parameter in time-series forecasting, often selected heuristically or through empirical tuning. Ermshaus et al.~\cite{ermshaus2023window} proposed an adaptive windowing approach based on seasonality heuristics to improve modeling in periodic time series. Hamon et al.~\cite{hamon2024unsupervised} evaluated different window lengths in an anomaly detection pipeline using TSFRESH but did not focus on predictive reliability or model stability. These contributions underscore the importance of sequence configuration but stop short of evaluating its effects on generalization or overfitting, particularly within BiLSTM-based forecasting. To address long-sequence modeling limitations in recurrent networks, recent DL architectures have introduced enhanced mechanisms for temporal representation. CMamba~\cite{arxiv2024a}, for example, leverages channel-wise correlation to support long-range dependency learning, while the Long Input Sequence Network (LISN)~\cite{arxiv2024b} uses structured pattern learning to improve generalization over extended sequences. Although these approaches offer valuable insights, they are not specifically focused on BiLSTM or robustness under noisy conditions in time-series forecasting.

Noise, often introduced by sensor errors or environmental variation, has been widely studied in classification tasks as a form of regularization~\cite{madry2018towards}, but its role in forecasting remains relatively less examined. Nazeri and Pisu~\cite{nazeri2023lstm} showed that additive noise can significantly degrade LSTM performance in microgrid prediction tasks. Similarly, Getu~\cite{getu2024lstm} observed model instability when forecasting over noisy signals. These findings highlight the need for a deeper understanding of how forecasting models behave under controlled noise conditions, especially in combination with sequence-related configurations.

Evaluation practices and preprocessing techniques also impact the reliability of model assessments. Kapoor and Narayanan~\cite{kapoor2023leakage} demonstrated that data transformations applied before temporal splitting can lead to test leakage and inflated performance metrics. Cerqueira et al.~\cite{cerqueira2020performance} recommend holdout-based evaluation over cross-validation for non-stationary time-series data to better preserve temporal integrity—an issue often overlooked in sequence-based models. Overfitting and limited generalization have also been associated with configuration-sensitive vulnerabilities. Hui et al.~\cite{hui2021privacy} found that overfitted models are more susceptible to membership inference attacks, particularly in small datasets. Yeom et al.~\cite{yeom2018privacy} and Chen et al.~\cite{chen2020machine} showed that poor generalization can increase exposure to extraction and evasion attempts. While this study does not simulate adversarial attacks, such findings support the broader goal of understanding how model configurations influence robustness.

Prior work has provided important insights into how sequence length, noise, and evaluation strategies individually influence forecasting performance. However, the combined effect of these factors on BiLSTM-based forecasting models—particularly in terms of robustness and generalization across diverse temporal regimes—remains insufficiently addressed. This study contributes a controlled empirical analysis to fill that gap.

\section{Methodology}
\label{sec:methodology}

To evaluate the impact of input sequence length and noise on the robustness of time-series forecasting models, we employ a BiLSTM architecture. This study examines how variations in these factors affect predictive stability across datasets with differing temporal resolutions. Controlled experiments are conducted under both clean and noisy input conditions to assess performance degradation and sensitivity to generalization. The objective is to inform data-centric evaluation strategies that enhance the reliability of DL-based forecasting pipelines.

\section{Dataset and Preprocessing}
\label{Dataset}

Weather forecasting plays a critical role in IoT applications such as smart agriculture, energy management, and disaster response, where time-series predictions must be both accurate and robust to noise. In these settings, data corruption or model failure can lead to unsafe or costly outcomes.

\begin{table}[!htbp]
    \centering
    \caption{Summary of the Datasets Used in the Study}
    \label{tab:dataset_summary}
    \renewcommand{\arraystretch}{1.0}
    \setlength{\tabcolsep}{3pt} 
    \scriptsize 
    \begin{tabular}{l l l c c c}
        \toprule
        \textbf{ID} & \textbf{Dataset} & \textbf{Target Variable} & \textbf{Samples} & \textbf{Time Span} & \textbf{Frequency} \\
        \midrule
        \textbf{D1} & Daily Climate Data & Mean Temperature ($^\circ$C) & \textbf{1,426} & 2013--2016 & Daily \\
        \textbf{D2} & Weather History & Temperature ($^\circ$C) & \textbf{95,936} & 2006--2016 & Hourly \\
        \textbf{D3} & Air Quality & Temperature ($^\circ$C) & \textbf{35,044} & 2013--2017 & Hourly \\
        \bottomrule
    \end{tabular}
\end{table}
This study employs three time-series datasets focused on weather forecasting, each using temperature as the target variable. The datasets vary in sample size, time span, and measurement frequency, as summarized in Table~\ref{tab:dataset_summary}. Specifically, we consider: daily climate records (D1), historical weather observations (D2), and air quality measurements (D3). This diversity enables a comprehensive evaluation of model performance across different temporal resolutions within a univariate forecasting setting.

Each dataset comprises timestamped observations structured as:
\[
\tau_i = \langle (t_1, x_1), \dots, (t_m, x_m) \rangle
\]
where \( x_k \) denotes the temperature recorded at time \( t_k \). Given the sequential nature of the data, it is essential to consider both noise levels and the risk of data leakage—particularly as a function of sequence length. Longer sequences can capture richer temporal dependencies but also risk overlapping between training and validation windows, potentially introducing correlation that leads to overfitting and inflated performance metrics. Such risks can undermine the reliability of performance evaluations, especially when overlapping sequences introduce information leakage between training and validation sets. To address this, we implement a systematic testing framework that incorporates robust statistical assessments and controlled noise-injection strategies, aiming to preserve data integrity and improve the generalization of forecasting models under varying temporal conditions. The key steps include:

\begin{itemize}
    \item Conversion of date columns into datetime format and setting them as the index to maintain temporal consistency.
    \item Normalization of the temperature variable using Min-Max
Scaling to prevent numerical instabilities and enhance model
robustness.
    \item Statistical noise assessment using the Ljung-Box and Aug-
mented Dickey-Fuller (ADF) tests to evaluate randomness
and stationarity in the dataset.
    \item Sliding window transformation to create overlapping se-
quences of past observations, enabling the BiLSTM model
to capture temporal dependencies.
    \item Evaluation of sequence length to assess its impact on data leakage and model generalization.
    \item Controlled Gaussian noise injection to simulate real-world
perturbations and assess model resilience against adversarial
and natural noise.
\end{itemize}


\subsection{Experimental Design}

To systematically investigate the impact of sequence length and noise injection on  time-series forecasting models, we design a series of controlled experiments focused on model robustness and reliability. These experiments are applied consistently across the datasets introduced in Section~\ref{Dataset}, enabling a comparative evaluation of how data characteristics influence model behavior under varying input configurations. The experimental procedure involves the following stages:

\begin{enumerate}
    \item \textbf{Real Dataset Evaluation:} Perform statistical tests, including the Ljung–Box white noise test and the Augmented Dickey–Fuller (ADF) test, to assess the randomness and stationarity of each dataset, ensuring their suitability for time-series modeling.

    \item \textbf{Baseline Model Training:} Train  model using a sequence length of 10, without any noise injection. This serves as a baseline for clean performance and model stability.

    \item  \textbf{Sequence Length Evaluation study:} We train models on extended sequences of length 30 to evaluate how increased temporal dependencies affect generalization, overfitting. This is compared against a baseline using a shorter sequence length of 10, enabling us to assess how input window size influences model robustness across datasets with varying sampling frequencies and sizes. 
    
   Although sequence length does not inherently cause data leakage, improper alignment of overlapping sequences during temporal train-test splitting can create correlated inputs across partitions. This may lead to the unintended inclusion of future information in the training set, resembling classical data leakage and inflating performance metrics. A unified partitioning strategy is applied across all datasets to isolate training from validation and testing phases, ensuring that the effect of sequence length is evaluated independently across datasets with varying sample sizes. While longer input sequences can capture richer temporal patterns, they also increase the risk of overfitting and reduce model generalization—particularly in data-scarce, low-frequency environments. By systematically analyzing sequence length across multiple datasets, we aim to better understand its impact on forecasting performance and its potential role in exposing models to robustness and configuration-sensitive risks.

\item \textbf{Noise Impact Study:} each model is trained on input sequences of length 10, with additive Gaussian noise applied to the target variable to assess its impact on forecasting accuracy and robustness. The noise is sampled from a normal distribution with a standard deviation of 0.05, simulating real-world uncertainty and enabling a systematic evaluation of the model's sensitivity to noisy outputs. This controlled perturbation strategy provides a means to examine model reliability under adverse conditions and to analyze its influence on forecasting stability.

\item \textbf{Interaction Study of Sequence Length and Noise:} Train models on sequences of length 30 with added Gaussian noise to evaluate how noise interacts with vulnerabilities introduced by extended sequence lengths, and how this interaction impacts model performance.

\begin{table}[!htbp]
    \centering
    \caption{BiLSTM Model Architecture and Hyperparameters}
    \label{tab:model_architecture}
    \renewcommand{\arraystretch}{1.0}
    \setlength{\tabcolsep}{4pt}
    \scriptsize
    \begin{tabular}{l l l}
        \toprule
        \textbf{Component} & \textbf{Hyperparameter} & \textbf{Value} \\
        \midrule
        \multirow{3}{*}{Recurrent Layers} 
            & BiLSTM layers & 3 \\
            & Hidden units & [128, 64, 32] \\
            & Activation & Tanh \\
        \midrule
        Regularization & Dropout rate & 0.2 \\
        \midrule
        \multirow{2}{*}{Optimization} 
            & Optimizer & AdamW \\
            & Learning rate & 0.001 \\
        \midrule
        Training & Early stopping & 10 epochs \\
        \bottomrule
    \end{tabular}
\end{table}

    \item \textbf{Cross-Experiment Evaluation:} Compare results across all conditions using standard performance metrics (e.g., RMSE, R\textsuperscript{2}), and analyze trends in performance degradation, generalization, and susceptibility to instability under noisy or extended sequence conditions.

\item \textbf{Comparative Analysis:} Evaluate all experimental conditions for each dataset using predefined performance metrics to assess differences in model behavior and robustness.

\item \textbf{Dataset-Level Replication and Comparison:} Repeat the full experimental procedure across all three datasets to observe how noise and sequence length affect forecasting performance within the same domain—weather sensor data—across datasets with varying sample sizes. This allows for a comparative analysis of model resilience under different data density conditions, offering insight into the generalizability of BiLSTM models under variable data regimes.


\end{enumerate}

This framework enables a structured and reproducible evaluation of time-series forecasting models, offering insights into how data quality and temporal structure influence model robustness.

\subsection{Model Architecture and Training}

A sequential BiLSTM model was implemented with an emphasis on balancing predictive accuracy and robustness. The architecture includes stacked recurrent layers, dropout regularization, and early stopping to mitigate overfitting. Detailed model components and hyperparameters are summarized in Table~\ref{tab:model_architecture}.

The dataset is partitioned using a three-way holdout strategy, allocating 70\% of the data for training, 15\% for validation, and 15\% for testing. This setup supports robust model evaluation while maintaining generalization performance.




\subsection{Performance Metrics}

To evaluate the statistical characteristics of the datasets and the predictive performance of the model, we employ a combination of statistical tests and forecasting metrics, as summarized in Table~\ref{tab:metrics}. This evaluation strategy supports rigorous data characterization and model assessment, enabling more reliable forecasting under noisy and dynamic conditions.

\begin{table}[!htbp]
    \centering
    \caption{Evaluation Metrics}
    \label{tab:metrics}
    \renewcommand{\arraystretch}{1.0}
    \setlength{\tabcolsep}{4pt}
    \scriptsize
    \begin{tabular}{l l l}
        \toprule
        \textbf{Metric} & \textbf{Category} & \textbf{Objective} \\
        \midrule
        White Noise Test & Statistical & Detect stochastic randomness \\
        Augmented Dickey-Fuller (ADF) & Statistical & Test for temporal stationarity \\
        RMSE & Predictive & Quantify error magnitude \\
        $R^2$ Score & Predictive & Measure explained variance \\
        \bottomrule
    \end{tabular}
\end{table}

\begin{table}[!htbp]
    \centering
    \caption{White Noise and Stationarity Test Results}
    \label{tab:statistical_tests}
    \renewcommand{\arraystretch}{0.95}
    \setlength{\tabcolsep}{4pt}
    \scriptsize
    \begin{tabular}{l c}
        \toprule
        \textbf{Test} & \textbf{Result} \\
        \midrule
        Ljung-Box Statistic (Lag = 10) & 12326.833 \\
        Ljung-Box p-value & 0.00001 \\
        ADF Statistic & -2.021 \\
        ADF p-value & 0.277 \\
        Critical Value (1\%) & -3.435 \\
        Critical Value (5\%) & -2.864 \\
        Critical Value (10\%) & -2.568 \\
        Stationarity Conclusion & Non-stationary ($p > 0.05$) \\
        \bottomrule
    \end{tabular}
\end{table}

\section{Results and Discussion}
\label{sec:results}

\subsection{Statistical Analysis of Data Quality}
To ensure the dataset's suitability, we conducted the Ljung-Box test to assess randomness and white noise characteristics, as well as the Augmented Dickey-Fuller (ADF) test to determine stationarity (See Table \ref{tab:statistical_tests}).

\paragraph{Ljung-Box Test (White Noise and Autocorrelation Analysis)} 
The Ljung-Box test, performed at lag 10, returned a p-value of \textbf{0.00001}, indicating that the time series is not purely random (white noise) and exhibits significant temporal dependencies. In this context, lag 10 signifies that the test evaluates autocorrelation up to 10 time steps in the past, determining whether past values influence future values over a 10-day window. This confirms that the dataset contains structured patterns, making it suitable for predictive modeling. The presence of autocorrelation supports the use of deep learning models, such as BiLSTM, which can effectively capture these dependencies for improved forecast accuracy.

\begin{table*}[t]
  \caption{Test Performance Metrics Across Datasets}
  \label{tab:test_performance_metrics}
  \centering
  \scriptsize 
  \resizebox{\textwidth}{!}{ 
  \begin{tabular}{lcccccc}
    \toprule
    \multirow{2}{*}{\textbf{Experiment Con.}} & \multicolumn{2}{c}{\textbf{D1}} & \multicolumn{2}{c}{\textbf{D2}} & \multicolumn{2}{c}{\textbf{D3}} \\
    \cmidrule(lr){2-3} \cmidrule(lr){4-5} \cmidrule(lr){6-7}
     & \textbf{RMSE} & \textbf{R²} & \textbf{RMSE} & \textbf{R²} & \textbf{RMSE} & \textbf{R²} \\
    \midrule
    Baseline & 0.049184 & 0.0949389 & 0.018132 & 0.986231 & 0.018800 & 0.991188 \\
    Extended Seq. Length & 0.058658 & 0.924205 & 0.017317 & 0.987482 & 0.018472 & 0.991256 \\
    Noisy & 0.093733 & 0.832568 & 0.084884 & 0.748299 & 0.086682 & 0.833420 \\
    Noisy + Ext. Seq. Length & 0.104970 & 0.784051 & 0.078842 & 0.784135 & 0.081879 & 0.846961 \\
    \bottomrule
  \end{tabular}}
\end{table*}

\paragraph{Augmented Dickey-Fuller (ADF) Test} 
The ADF test yielded a statistic of \textbf{-2.021} with a p-value of \textbf{0.277}, which exceeds the 5\% significance threshold. Since the ADF statistic is greater than the 5\% critical value (\textbf{-2.864}), we fail to reject the null hypothesis, indicating that the dataset is non-stationary. This suggests that the time series exhibits trends or seasonality, but given the strong predictive performance observed with non-stationary data, no additional transformations were applied.

\paragraph{Implications for Forecasting} 
As illustrated in Table \ref{tab:statistical_tests}, the results indicate that the original dataset is non-stationary, meaning that statistical properties such as mean and variance change over time. However, despite non-stationarity, the data retains strong autocorrelation, as confirmed by the Ljung-Box test. This suggests that past values provide meaningful information for forecasting, making deep learning models such as BiLSTM well-suited for capturing the underlying temporal dependencies and trends in the data. 

The evaluation of the experimental conditions across all three datasets reveals distinct patterns in model performance under varying sequence lengths and noise configurations. Table~\ref{tab:test_performance_metrics} summarizes the test RMSE and R\textsuperscript{2} scores for each condition. The results show that both longer sequences and noise injection independently affect model performance, while their combination leads to the most pronounced degradation—particularly in the D1 dataset, which has the smallest sample size and lowest temporal resolution.

For D1, RMSE increased from 0.049 in the baseline to 0.058 with extended sequences, 0.094 with noise, and peaked at 0.105 when both were combined. Its R\textsuperscript{2} dropped sharply from 0.095 in the baseline to 0.784 in the combined setting, highlighting sensitivity to noise and temporal extension. In contrast, D2 and D3, with higher-frequency (hourly) records and larger sample sizes, exhibited greater robustness. D2 achieved the lowest RMSE (0.017) and highest R\textsuperscript{2} (0.987) under extended sequences, even slightly outperforming the baseline. However, the addition of noise in D2 and D3 still caused substantial performance drops, particularly when combined with longer sequences.

\begin{figure}[h]
  \centering
  \includegraphics[width=0.6\linewidth]{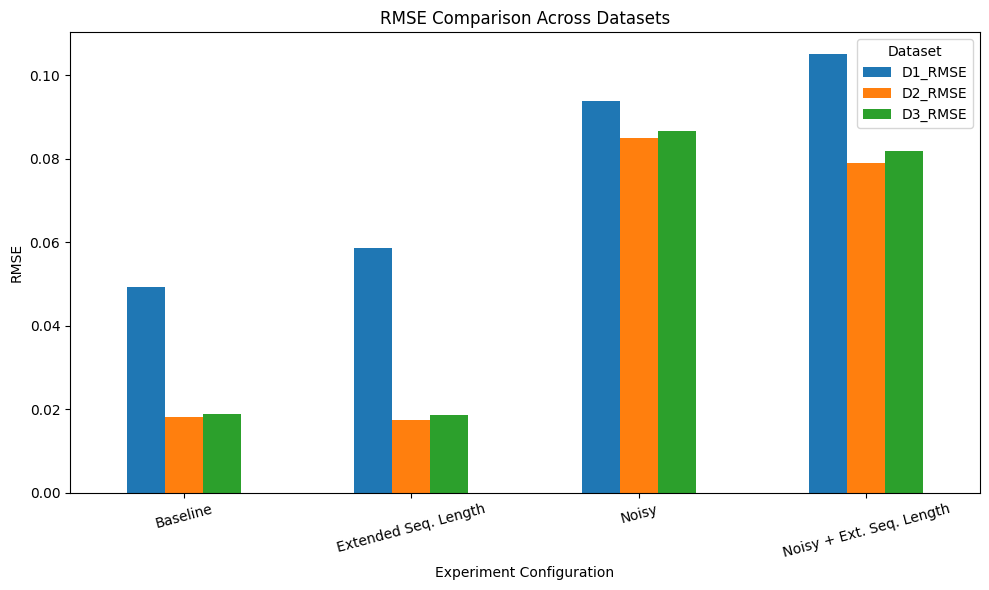} 
  \caption{RMSE Across Experimental Conditions}
  \label{fig:RMSE_metrics}
\end{figure}

These findings, visualized in Figure~\ref{fig:RMSE_metrics} and Figure~\ref{fig:R2_metrics}, underscore the importance of evaluating the interaction between sequence length and noise within a testing framework that considers their impact on model robustness and reliability. The results suggest that while high-frequency datasets benefit from deeper temporal modeling, their resilience diminishes under noisy conditions—especially when long input windows are used. This highlights the need for evaluation practices tailored to dataset characteristics that influence model stability and potential vulnerability.

\begin{itemize}
    \item \textbf{The Baseline model} achieves the best overall performance, with the lowest test RMSE across all datasets—\textbf{0.0492} (D1), \textbf{0.0181} (D2), and \textbf{0.0188} (D3)—and consistently high R\textsuperscript{2} scores. This confirms that a short sequence length of 10 enables the model to generalize effectively without overfitting, particularly in data-rich, high-frequency environments. This indicates that a moderate sequence length facilitates better generalization, whereas excessively long sequences may introduce unnecessary complexity without improving predictive accuracy.

\item  \textbf{The Extended Sequence Length} model, which uses a sequence length of 40, exhibits mixed results. While RMSE slightly improves in D2 and D3, it increases in D1 (\textbf{0.0587}), accompanied by a drop in R\textsuperscript{2} to \textbf{0.9242}. These findings suggest that longer sequences may capture richer temporal dependencies in large, high-frequency datasets but can lead to overfitting and reduced stability in smaller, low-frequency datasets. This behavior reinforces the hypothesis that increasing sequence length beyond a certain threshold may degrade generalization by encouraging the model to memorize patterns rather than learn transferable structures. As a result, such configurations may reduce model reliability, particularly in resource-constrained settings.

    \item  \textbf{The Noisy model}, which uses a short sequence (length 10) with Gaussian noise injection, demonstrates significant performance degradation in all datasets. RMSE increases to \textbf{0.0937} (D1), \textbf{0.0849} (D2), and \textbf{0.0867} (D3), accompanied by substantial drops in R\textsuperscript{2}. This highlights the dual role of noise: while it can obscure learned patterns to enhance robustness, it also introduces vulnerability by degrading predictive accuracy and potentially amplifying instability under adversarial conditions.

\item  \textbf{The Noisy Extended Sequence Length model} yields the weakest performance, with the highest RMSE—\textbf{0.1050} (D1), \textbf{0.0788} (D2), and \textbf{0.0819} (D3)—and the lowest R\textsuperscript{2} scores across all datasets. This outcome reflects the compounded negative effects of extended temporal dependencies and noisy inputs, where overfitting and input perturbations jointly degrade the model’s ability to generalize and maintain stability.

Our findings indicate that noise does not mitigate the impact of extended sequences. Instead, the two factors reinforce each other, amplifying performance degradation. As shown in Figure~\ref{fig:RMSE_metrics}, RMSE increases substantially under noisy conditions, with the highest values occurring when noise and long sequences are combined. Similarly, in Figure~\ref{fig:R2_metrics}, R\textsuperscript{2} values drop most significantly in this setting, confirming that longer sequences do not compensate for the instability introduced by noise but instead exacerbate it.

\begin{figure}[h]
  \centering
  \includegraphics[width=0.6\linewidth]{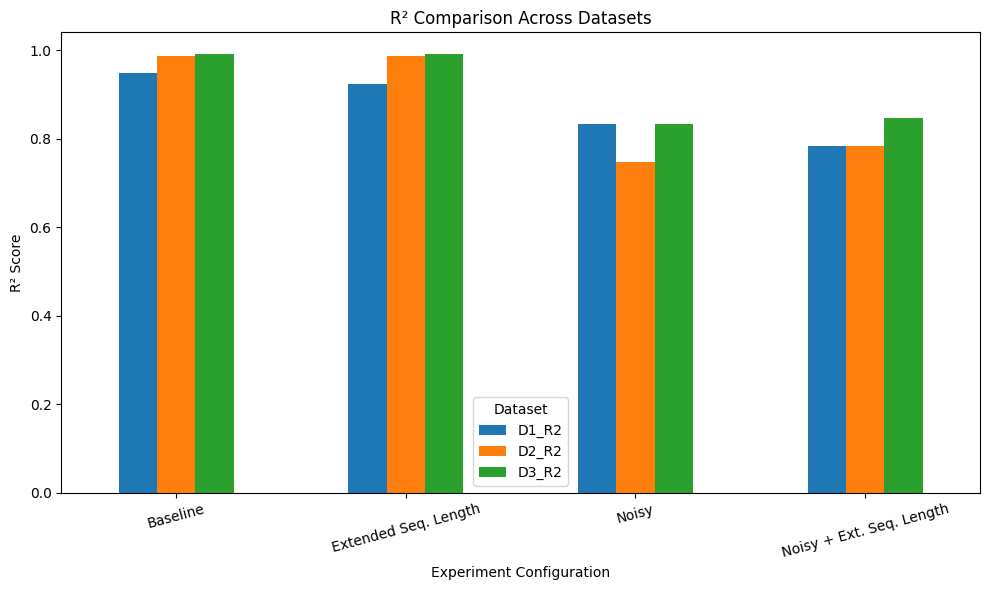}
  \caption{R\textsuperscript{2} Across Experimental Conditions}
  \label{fig:R2_metrics}

\end{figure}

This trend is particularly evident in D1, where the smallest dataset with the lowest temporal resolution exhibits the steepest decline in performance. In contrast, D2 and D3, which have larger sample sizes and higher temporal resolutions, demonstrate greater resilience but still suffer noticeable degradation under noise injection. These results suggest that while extended sequences capture richer temporal dependencies, they also introduce overfitting risks that are magnified in the presence of noise. These findings underscore the importance of carefully balancing sequence length and noise levels.

\end{itemize}


\subsection{Performance Under Varying Input Conditions}

\begin{figure}[t]
  \centering
  \begin{subfigure}[b]{0.48\linewidth}
    \centering
    \includegraphics[width=\linewidth]{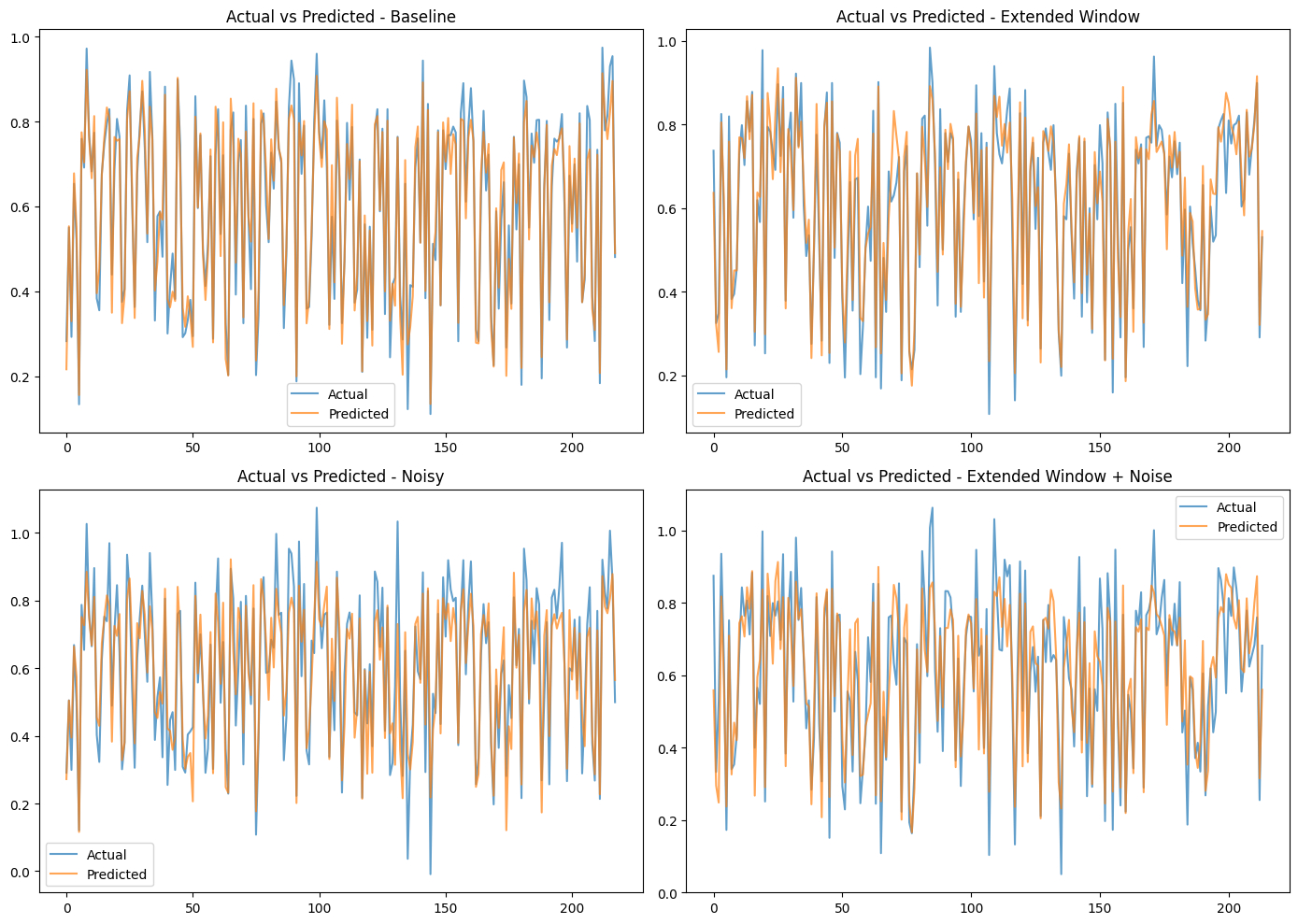}
    \caption{Actual vs. predicted values.}
    \label{fig:actual_vs_predicted_d1}
  \end{subfigure}
  \hfill
  \begin{subfigure}[b]{0.48\linewidth}
    \centering
    \includegraphics[width=\linewidth]{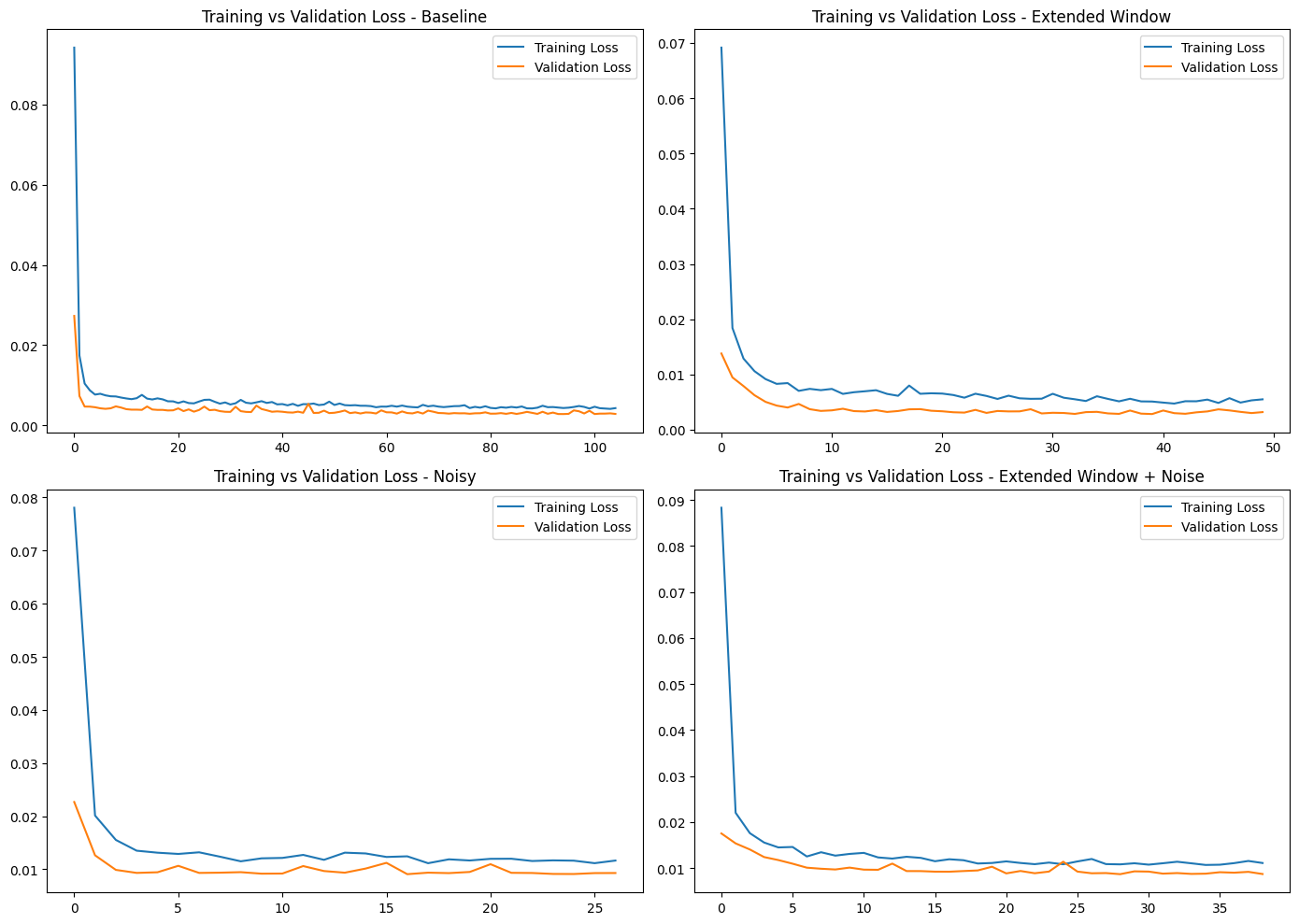}
    \caption{Training and validation loss.}
    \label{fig:training_vs_validation_d1}
  \end{subfigure}
  \caption{Forecasting results and loss curves for dataset D1.}
  \label{fig:d1_combined}
\end{figure}

\begin{figure}[t]
  \centering
  \begin{subfigure}[b]{0.48\linewidth}
    \centering
    \includegraphics[width=\linewidth]{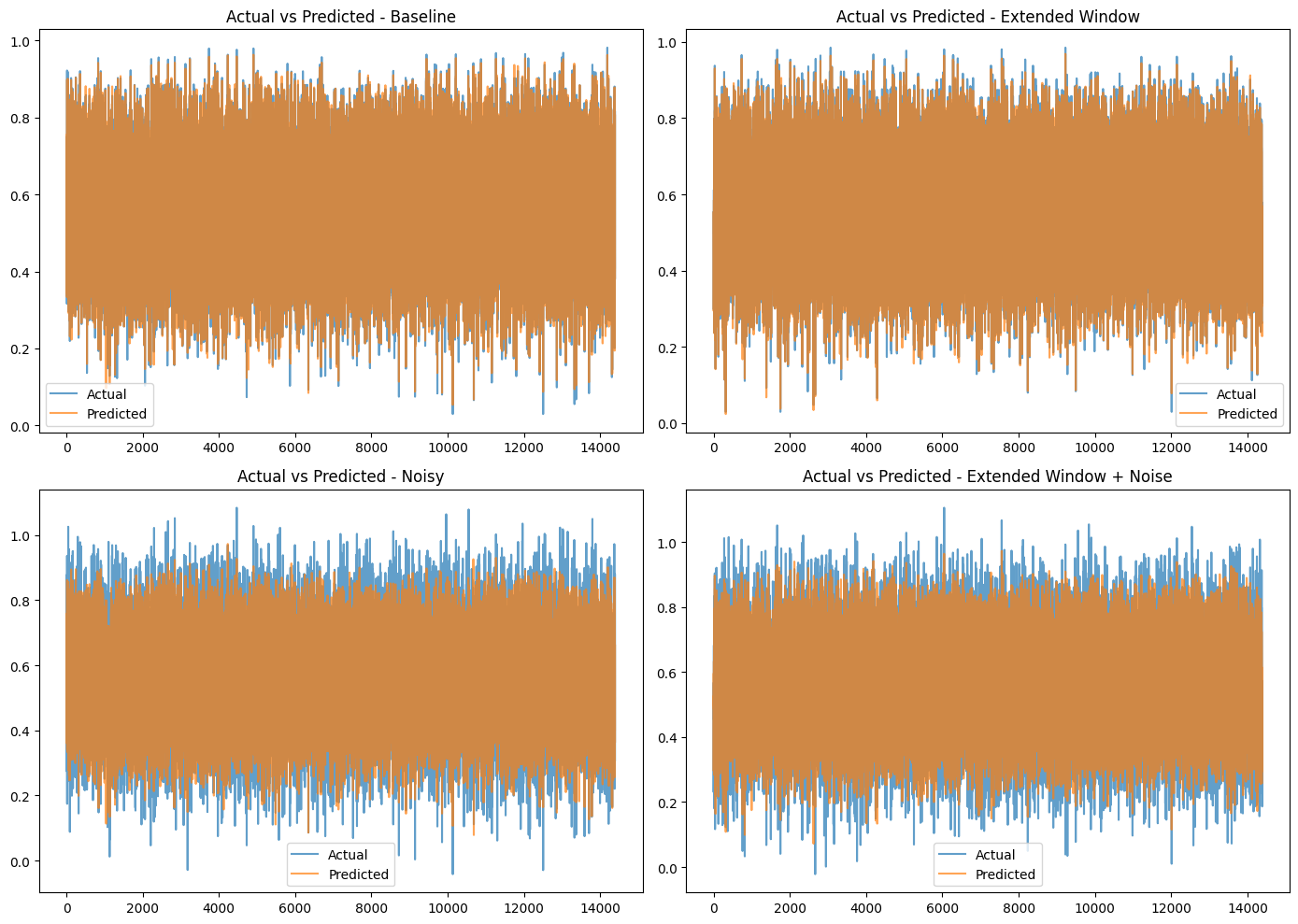}
    \caption{Actual vs. predicted values.}
    \label{fig:actual_vs_predicted_d2}
  \end{subfigure}
  \hfill
  \begin{subfigure}[b]{0.48\linewidth}
    \centering
    \includegraphics[width=\linewidth]{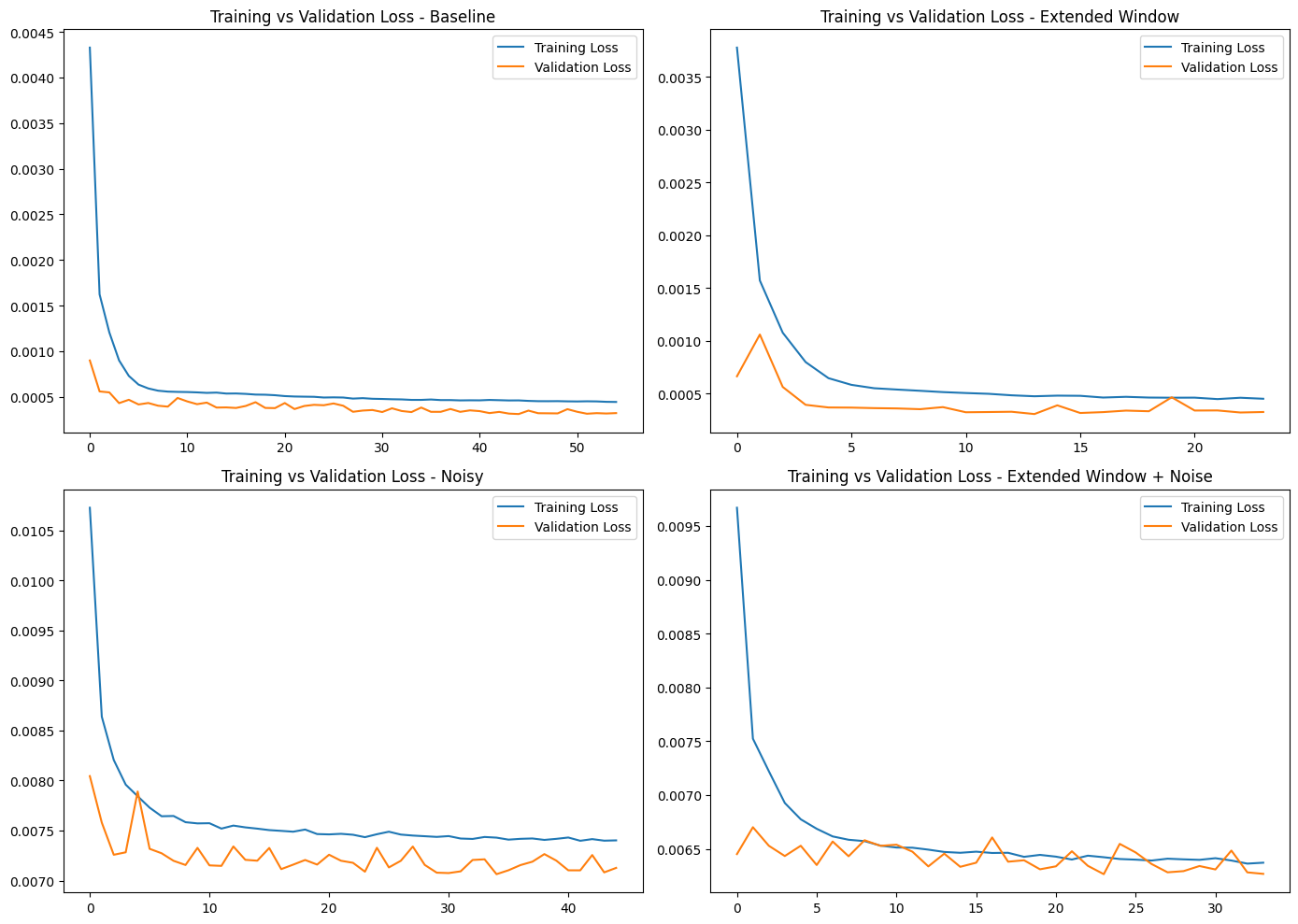}
    \caption{Training and validation loss.}
    \label{fig:training_vs_validation_d2}
  \end{subfigure}
  \caption{Forecasting results and loss curves for dataset D2.}
  \label{fig:d2_combined}
\end{figure}

The experimental results illustrate clear trade-offs between sequence length, noise injection, and forecasting performance. As shown in Figures~\ref{fig:actual_vs_predicted_d1} and~\ref{fig:actual_vs_predicted_d2}, the baseline model consistently achieves the closest alignment between actual and predicted values across all datasets, reflecting superior predictive accuracy and stability. In contrast, models trained with extended sequences or injected noise exhibit increased deviations, with the most pronounced discrepancies observed in D1. This suggests that smaller, lower-resolution datasets are more vulnerable to modeling instability.

Further evidence is provided in Figure~\ref{fig:training_vs_validation_d1}, where the extended-sequence model displays a widening gap between training and validation losses—particularly for D1—indicating a higher risk of overfitting. Although D2 and D3, characterized by higher sampling frequencies and larger sample sizes, demonstrate greater resilience, minor performance fluctuations still occur under extended sequence configurations (Figure~\ref{fig:training_vs_validation_d2}). Given space constraints and the similarity of trends in D2 and D3, we focus visual analysis on D1 and D2.

These results underscore the value of data-aware evaluation strategies. While longer sequences can capture richer temporal dependencies—particularly in high-frequency, data-rich settings—they also increase the risk of overfitting in smaller or coarser-grained datasets. Similarly, noise injection can enhance robustness by discouraging memorization and promoting generalization, aligning with prior work on regularization in deep learning. However, when noise is not properly calibrated to the data distribution, it can significantly degrade predictive accuracy.

Overall, these findings highlight the delicate balance between accuracy and stability in time-series forecasting. The compounded effects of long sequences and noise tend to exacerbate performance degradation in low-resolution or data-scarce scenarios. This reinforces the necessity of dataset-specific modeling choices, where input configurations—such as sequence length and noise level—are carefully tuned to the dataset's temporal resolution, sample size, and sensitivity to perturbations. Adopting such data-oriented strategies is essential not only for achieving optimal performance but also for ensuring generalization and reliability, particularly in high-stakes or resource-constrained forecasting applications.

\section{Future Work and Research Directions}
\label{sec:future_work}

This study establishes a foundation for data-centric testing in deep learning-based time-series forecasting. Future research will extend this framework to address challenges in generalization and robustness—key concerns in the development of reliable AI systems. Several promising directions include:

\begin{itemize}
    \item Developing adaptive sequence configuration algorithms that optimize forecasting accuracy across heterogeneous temporal resolutions.
    \item Investigating the interaction between preprocessing methods (e.g., normalization, detrending) and model behavior under noise, imbalance, and missing data.
    \item Expanding dataset-oriented testing to support complex time-series modalities, such as multivariate, event-based, or irregularly sampled data, relevant for healthcare and control systems.
    \item Studying model robustness under distributional shifts, including adversarial and synthetic perturbations, to enhance reliability in real-world deployment scenarios.
    \item Integrating explainable AI techniques to interpret sequence sensitivity and noise tolerance in recurrent and attention-based architectures.
    \item Designing reproducible, lightweight benchmarks for evaluating forecasting models at scale, with a focus on responsible AI practices in high-stakes domains such as finance, energy, and healthcare.
\end{itemize}

By aligning time-series evaluation with emerging directions in explainability, robustness, and responsible AI, this research contributes toward building transparent and dependable forecasting systems across scientific and industrial applications.

\section{Conclusion }  
\label{sec:conclusion}

This study investigates the integration of data-oriented testing within deep learning for time-series forecasting, with a focus on model configurations that influence robustness and reliability in sensitive deployment contexts. By analyzing the impact of sequence length and noise injection on BiLSTM-based models across three datasets with varying sample sizes and temporal resolutions, we highlight the critical interplay between data characteristics and model behavior. Our findings reveal that longer sequence lengths can improve temporal representation but also increase the risk of overfitting, particularly in low-frequency datasets. This leads to reduced generalization and greater model instability. Noise injection, while capable of disrupting memorized patterns, consistently degrades predictive accuracy—especially when combined with extended sequences. These results underscore the importance of dataset-aware configuration, where sequence length and noise must be carefully balanced to prevent performance degradation and support stable forecasting.
Testing plays a vital role in evaluating how configuration choices affect model robustness, providing insights that are relevant to secure software testing and trustworthy ML pipeline design. Our work demonstrates that sequence length and noise injection have dataset-dependent effects, reinforcing the need for systematic testing approaches in time-series model development. Future work should explore adaptive tuning strategies and testing procedures that improve robustness without compromising predictive accuracy. Such practices will be essential for building dependable forecasting systems in critical applications.


\begin{credits}
\subsubsection{\ackname} The authors would like to acknowledge the support received from the Saudi Data and AI Authority (SDAIA) and King Fahd University of Petroleum and Minerals (KFUPM) under the SDAIA-KFUPM Joint Research Center for Artificial Intelligence Grant JRC-AI-RFP-20.

\subsubsection{\discintname}
 The authors have no competing interests to declare that are relevant to the content of this article.

\end{credits}
%
%
%
\bibliographystyle{splncs04}
\bibliography{Ref}
%




\end{document}